\documentclass[lettersize,journal]{IEEEtran}
\usepackage{amsmath,amsfonts}
\usepackage{algorithmic}
\usepackage{algorithm}
\usepackage{array}
\usepackage[caption=false,font=normalsize,labelfont=sf,textfont=sf]{subfig}
\usepackage{textcomp}
\usepackage{stfloats}
\usepackage{url}
\usepackage{verbatim}
\usepackage{graphicx}
\usepackage{cite}
\usepackage{multirow}
\usepackage{bigstrut}
\usepackage{colortbl}
\usepackage{xcolor}
\usepackage{caption}
\usepackage{overpic}
\usepackage{booktabs}

\usepackage[capitalize]{cleveref}
\crefname{section}{Sec.}{Secs.}
\Crefname{section}{Section}{Sections}
\Crefname{table}{Table}{Tables}
\crefname{table}{Tab.}{Tabs.}

\usepackage{soul}
\graphicspath{{figures/}}

\begin{document}

\title{Multi-Content Interaction Network for Few-Shot Segmentation}

\author{Hao Chen, Yunlong Yu, Yonghan Dong, Zheming Lu, ~\IEEEmembership{Senior Member,~IEEE}, Yingming Li, and Zhongfei Zhang, ~\IEEEmembership{Fellow,~IEEE}
\thanks{Hao Chen, Yunlong Yu, Zheming Lu, and Yingming Li are with Zhejiang University, Hangzhou, 310027, P.R. China. E-mail: {chen\_hao\_zju, yuyunlong, zheminglu, yingming}@zju.edu.cn }
\thanks{Yonghan Dong is with Huawei Technologies Ltd. E-mail: dongyonghan@huawei.com }
\thanks{Zhongfei Zhang is with State University of New York, Binghamton, NY 13902-6000 USA. E-mail: zhongfei@cs.binghamton.edu }
}

\markboth{Journal of \LaTeX\ Class Files,~Vol.~14, No.~8, August~2021}%
{Shell \MakeLowercase{\textit{et al.}}: Multi-Content Interaction Network for Few-Shot Segmentation}


\maketitle

\begin{abstract}
Few-Shot Segmentation (FSS) is challenging for limited support images and large intra-class appearance discrepancies. Most existing approaches focus on extracting high-level representations of the same layers for support-query correlations, neglecting the shift issue between different layers and scales, due to the huge difference between support and query samples. In this paper, we propose a Multi-Content Interaction Network (MCINet) to remedy this issue by fully exploiting and interacting with the multi-scale contextual information contained in the support-query pairs to supplement the same-layer correlations. Specifically, MCINet improves FSS from the perspectives of boosting the query representations by incorporating the low-level structural information from another query branch into the high-level semantic features, enhancing the support-query correlations by exploiting both the same-layer and adjacent-layer features, and refining the predicted results by a multi-scale mask prediction strategy, with which the different scale contents have bidirectionally interacted. Experiments on two benchmarks demonstrate that our approach reaches SOTA performances and outperforms the best competitors with many desirable advantages, especially on the challenging COCO dataset.  
\end{abstract}


\section{Introduction}
\label{sec:intro}

\IEEEPARstart{W}{ith} the rapid development of computer vision, semantic segmentation \cite{fcn, u-net, semantic_fpn}, as one of the most important vision fields has achieved remarkable performances. Such great success is partly attributed to the availability of a large amount of labeled training data. However, in practice, obtaining pixel-level image annotations is especially tedious and expensive. To alleviate the annotating efforts, Few-Shot Segmentation (FSS) \cite{oslsm, dan, panet, sgone} which segments the target samples with only one or a few support samples, has been attracting a lot of attention.

\begin{figure}[t]
	\centering
	\subfloat[Existing Class-wise Framework]{\includegraphics[width=3in]{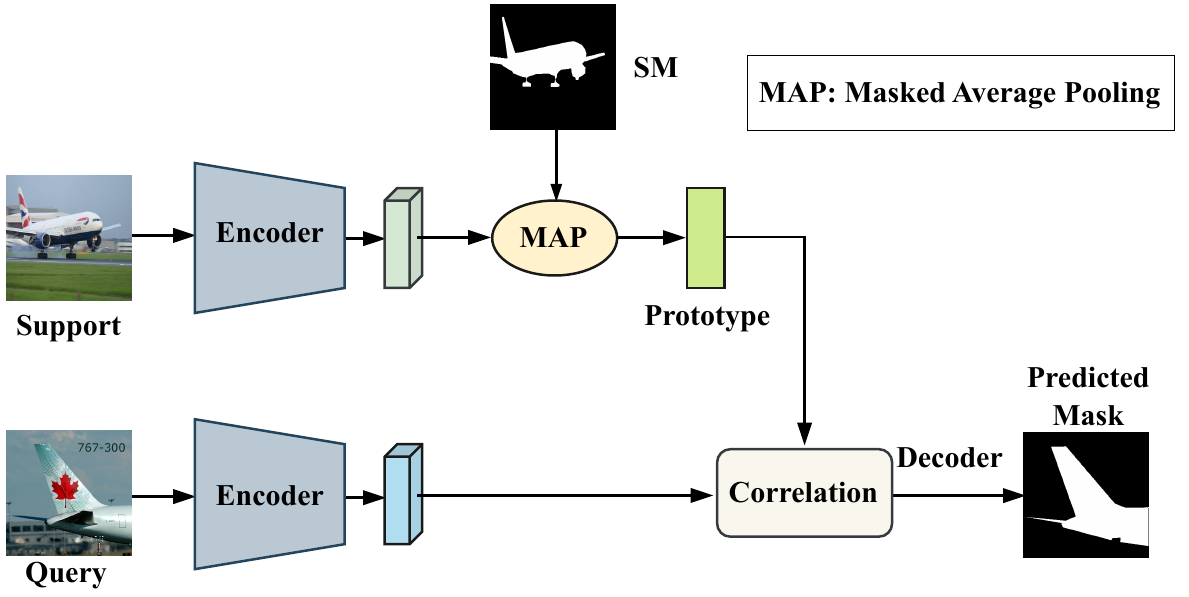}%
		\label{fig_a}}
	\hfil
	\vspace{2mm}
	\subfloat[Our Framework]{\includegraphics[width=3in]{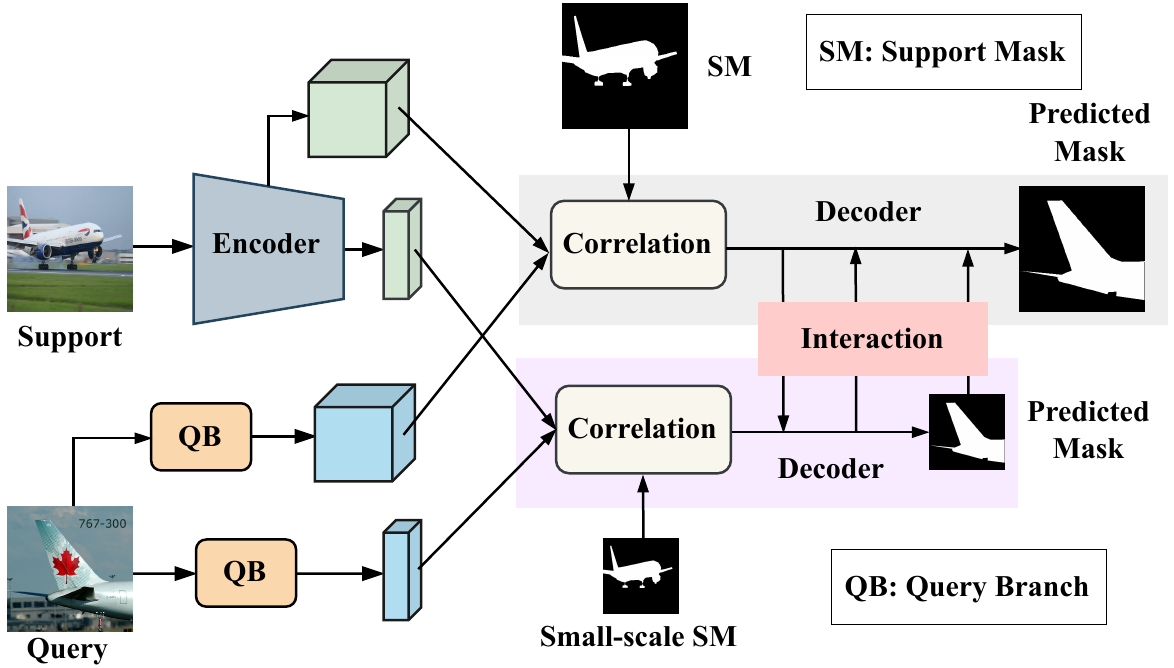}%
		\label{fig_b}}
	\caption{Comparison between the existing FSS framework and ours(MCINet) for the 1-Shot segmentation task. The main difference is that the former considers the support-query correlations with a class-wise prototype, while we densely and comprehensively consider pixel-wise relationships, including exploiting and interacting multi-scale information.}
	\label{fig:sketch}
\end{figure}

FSS remains a challenging task due to that the models hardly capture the large intra-category appearance variations with only a small support set composed of only one or a few images. A key to FSS is to fully exploit the information contained in the support images that could distinguish the target objects in the query images from the backgrounds. Most of the existing FSS approaches \cite{panet, pfenet, bam, ntrenet} make full use of the limited support images via abstracting the support objects into a class-wise prototype with the masked average pooling \cite{sgone}, against which the query features were predicted the object masks of the query images, as shown in \cref{fig:sketch} \subref{fig_a}. As the average pooling way of obtaining the class prototype inevitably results in a great information loss contained in the support images, it is suboptimal to apply the class-wise support prototype to predict the query masks. 

To address this issue, some recent works \cite{hsnet, vat} explore the pixel-wise correlations between the query images and the foreground support features and have shown some advantages against the prototype-based approaches. However, these approaches ignore the backgrounds of the support images. \cite{cyctr, dcama} further take the support backgrounds into consideration and fully investigate the dense correlations between the support and the query images. Though promising performances have been achieved, the existing approaches focus on the same-layer correlation of query-support pairs, neglecting the shift issue, where the most related support-query features may be not in the same layer, due to the huge difference between query and support samples. Besides, the existing approaches only focus on the high-level information contained in the backbone while the low-level structural information does not consider enough, which is important for the pixel-level segmentation tasks. 

Thus, we argue that wide content interaction between support-query pairs benefits extracting the pixel-wise correlations due to the huge difference between support and query samples. In this work, we propose a method called Multi-Content Interaction Network (MCINet) to fully exploit and interact with the multi-scale and multi-layer information of different content contained in the support-query pairs. Specifically, MCINet improves FSS from the perspectives of enhancing the query feature representations, the support-query correlations, and the query mask predictions. 



Considering the important role of low-level structural information plays in pixel-level prediction tasks, Our method focuses on boosting the query feature representations via incorporating the low-level information with the different content. Though many methods explore the supplementary effect of multi-scale features from the same backbone, where the low-level and high-level features are highly related, we fuse cross-scale features from the different branches. The unrelated content from the different branches will alleviate the low-level information loss via replenishing the structural information to the high-level features. Specifically, we first transform the input query images into the same size as the high-level feature maps with a downsampling and a simple convolutional operation and then apply a cross-attention module to fuse the low-level information into the high-level representations.


The existing approaches focus on support-query correspondence from the same layer of backbone, which neglects the shift issue between different layers and scales. We present a novel strategy to exploit the support-query affinity from different scales, concerning the adjacent-layer similarity to remedy the inconsistency between adjacent layers. 


During the mask prediction process, the existing approaches train the model under only the supervision of the original input-size masks. However, for dense segmentation tasks, especially for large-scale semantic segmentation, the intermediate modules are vital for the final predictions. If no supervision is imposed on the intermediate modules of the deep networks, they might embarrass the final mask predictions. To this end, we present to segment the query masks under the supervision of multi-scale ground-truth masks, where each scale enjoys a decoder branch. With this multi-scale mask prediction strategy, the information of different scales could be fully interacted in a bidirectional way. The basic differences between the existing prototype-based approaches and ours are shown in \cref{fig:sketch}. Though the multi-scale works, including the multi-scale transformer \cite{chen2016attention, gu2022multi} and multi-scale fusion \cite{upernet, deeplabv3+}, have been extensively explored in the generic semantic segmentation, the multi-scale mask prediction strategy is rarely studied for FSS. Different from feature extraction like the existing Semantic Segmentation methods, our multi-scale supervision facilitates the model in learning support-query affinity.






In conclusion, the contributions of this work include:

\begin{itemize}
    \item We introduce a Multi-Content Interaction Network (MCINet) for FSS, which supplements the multi-content correlations from different layers of support-query pairs during the feedforward network by exploiting and interacting with the feature from another query branch for support-query pairs.


    \item We first introduce adjacent-layer similarity to remedy the shift issue between query and support features, and enhance the support-query correlations from same-scale and adjacent-scale. 
    
    \item As far as we know, we are the first to segment the query images with a multi-scale mask prediction strategy for learning support-query affinity, which bidirectionally interacts with the information among different scales. 
\end{itemize}


    

Extensive experiments show that our approach sets new SOTAs on two FSS benchmarks and outperforms the competitors with many desirable advantages, especially on the challenging COCO dataset \cite{lin2014microsoft}. In addition, we also conduct an ablation study and provide visualizations to validate our MCINet.

\section{Related work}
\label{sec:related}

\subsection{Few-Shot Learning}
As deep learning evolves, researchers think about how to make full use of every sample with scarce labeled samples. One solution is Few-Shot Learning (FSL), which attempts at learning novel concepts with the guidance of scarce annotated samples. The earlier FSL works mainly address classification \cite{matching-network, maml, mm-net} and recently have been widely extended to object detection \cite{fsce, fsod, fsrw, repmet} and semantic segmentation \cite{oslsm, ppnet, pfenet, repri, cwt}. The most popular FSL pipeline is based on meta-learning \cite{metanets}, which is divided into three main streams, metric-based \cite{prototypical, tadam, tapnet, ctm}, optimization-based \cite{lstm-meta, mtl}, and parameter-based \cite{csn}.

\begin{figure*}[t]
  \centering
   \includegraphics[width=0.98\linewidth]{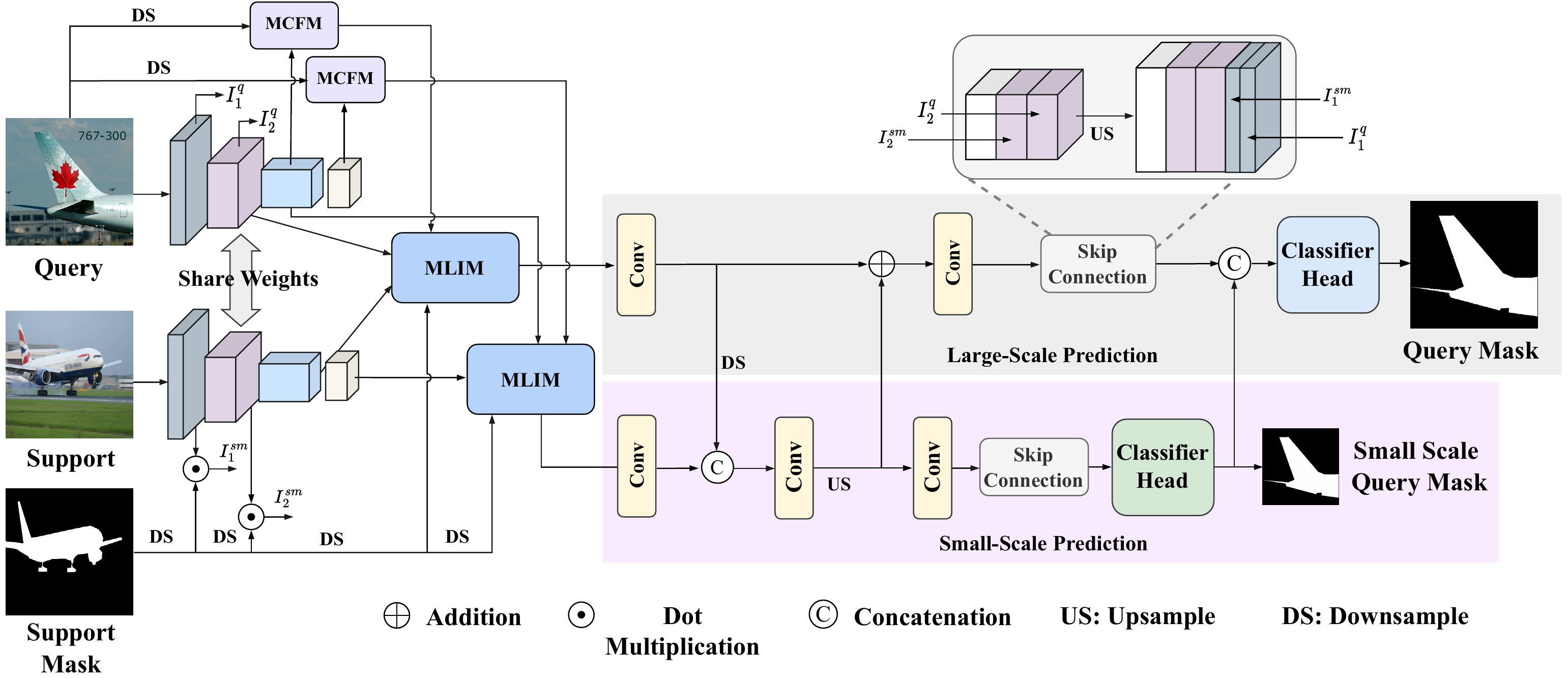}
   \caption{The framework of the proposed MCINet. MCINet takes the features of different scales enhanced by the low-level structural information with a Multi-Content Fusion Module (MCFM, please refer to \cref{mlfm}) for the support-query correlation. The support-query correlation is obtained with a Multi-Layer Interaction Module (MLIM, please refer to more details in \cref{msim}). The query image is segmented with a multi-scale mask prediction strategy, where the small-scale mask predictions assist in the mask predictions of the target large-scale images. `FC' is short for fully connected layer. `Conv' is short for convolutional layer. }
   \label{fig:framework}
\end{figure*}

\subsection{Few-Shot Segmentation}
The most popular FSS pipeline is based on the metric learning, which segments the query images by computing the correlation between the support prototype and each pixel representation of the query image. 
 Most existing approaches vary in the way of obtaining the support prototypes. As one of the earliest works to adopt prototype learning in FSS, SGONE \cite{sgone} uses the masked average pooling (MAP) to extract the support prototype. To extract more representative support prototypes, ASGNet \cite{asgnet} extracts multiple prototypes and picks the most representative one, DPCN \cite{dpcn} adopts a dynamic prototype to fully capture discontinuous mask, Ntrenet \cite{ntrenet} extracts a background prototype to eliminate the similar region in the query image, while DCP \cite{dcp} does a more refined work that extracts two prototypes and computes correlation with query image in foreground and background separately. Though prototype learning has been extensively explored, it is still challenging to extract a representative support prototype to represent the whole class with a few support images.

Recently, the pipeline that directly computes the pixel-wise correlations between the support-query pairs obtains very competitive performances in FSS. DAN \cite{dan} is one of the earliest works to compute pixel-wise correlation in FSS, which enhances query-support correlation through graph attention mechanism. Following the similar idea, HSNet \cite{hsnet} computes the pixel-wise correlation between support-query pairs and enhances the correlation matrix with a 4D convolutional operation. VAT \cite{vat} extends the correlation enhancement module from a 4D convolutional network to a 4D swin transformer \cite{liu2021swin}. Though the pixel-wise correlation could retain the most abundant category information, these approaches might result in unnecessary information loss as they ignore the support background. 

DCAMA \cite{dcama} remedies this issue by considering the support background with a dense pixel-wise attention-weighted mask aggregation module. In this work, we also consider the support background to obtain the pixel-wise support-query correlation. Differently, our approach enhances the support-query correlation by fully exploiting multi-scale information fused with a 4D convolutional operation, promoting the foreground-background interaction between support-query pairs.

\section{Methods}
\label{sec:Method}

\subsection{Problem Definition}

Given only one or a few support images, FSS aims to segment objects of target classes with the model trained from the base classes. Note that the base classes and the target classes are disjoint in the label space. In this work, we follow the mainstream FSS approaches and apply a meta-learning paradigm \cite{oslsm} to train the model with a collection of episodes sampled from the base classes, where each episode simulates an FSS task that consists of a support set $\mathcal{S}={\{ I_{k}^{s},\ M_{k}^{s}\}}_{k = 1}^{K}$ and a query set $\mathcal{Q} = \{ I^{q} \}$, $I$ represents the input support image, $M$ denotes the corresponding mask, and $K$ denotes the number of support images. The model $H$ is trained with the objective of predicting the masks of the query images with the support set, \textit{i.e.}, $M^{q} = H(I^{q},\ {\{ I_{k}^{s},\ M_{k}^{s}\}}_{k = 1}^{K})$.

Following \cite{dcama,min2021hypercorrelation}, we exploit both the intermediate layer and the final layer features to address FSS. Specifically, we input both the support and query images to a pre-trained backbone to respectively extract the collection of their multi-scale multi-layer feature maps $\{F_{i,l}^s\}$ and $\{F_{i,l}^q\}$, where $i$ is the scale of the feature maps and $l \in \{1, ..., L_i\}$ is the index of all layers in a specific scale $i$. In this work, we only take the last two scale feature maps as the feature representations. Note that the feature extractor backbone is frozen during training and we design models on the top of the backbone to enable the whole network to segment the query images.

As illustrated \cref{fig:framework}, we improve FSS from the perspectives of fully exploiting and comprehensively interacting with the multi-scale information from the support-query pairs, which includes enhancing the query feature representations with a Multi-Content Fusion Module (MCFM), exploiting the support-query correlations with a Multi-Layer Interaction Module (MLIM), and improving the mask prediction with a Multi-Scale Mask Prediction (MSMP) strategy. In the following, we take 1-Shot segmentation as an example to introduce them in detail. We follow \cite{dcama} to expand the 1-Shot task to the 5-Shot segmentation task.

\subsection{Multi-Content Fusion Module}
\label{mlfm}

Enhancing the feature representations with the attention modules has been widely explored \cite{cyctr,catrans}. However, these works focus on improving the high-level semantic information or fusing the low-level and high-level features from the same backbone while the highly related features lack uniqueness. To remedy this, we present a simple Multi-Content Fusion Module (MCFM) to fuse the low-level structural information from another query branch into the high-level semantic features for the query images. 

\begin{figure}[t]
  \centering
  \includegraphics[width=1\linewidth]{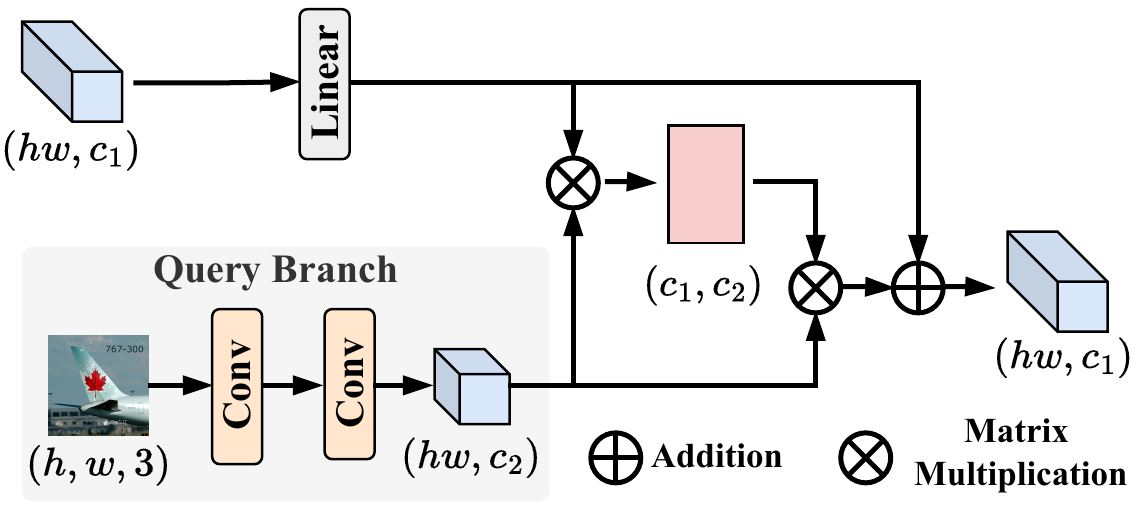}
  \caption{Illustration of Multi-Content Fusion Module (MCFM) under the 1-Shot segmentation task.}
  \label{fig:csfm}
\end{figure}

As illustrated in \cref{fig:csfm}, MCFM is conducted on the query feature extraction, which consists of a down-sampling operation, a query branch containing two convolutional layers and a cross-attention module. Specifically, given a query image $I^{q} \in \mathbb{R}^{3 \times H \times W}$, $H$ and $W$ denote the height and width, respectively, we first downsample the query image with the bilinear interpolation to reduce the spatial scale to the same size of the $i$-layer high-level feature map and boost the channel numbers of the query image from 3 to 128 with two convolutional layers, which is formulated as: 
\begin{equation}
    \mathbf{I}_{i}^{q} = f_{2\times conv}(DS(I^{q}) \downarrow),
\end{equation}
where $DS$ and $f_{2\times conv}$ denote the dowsampling operation and two convolutional layer, respectively.  

Due to the channel differences between the low-level and the high-level representations, we apply a cross-attention module to fuse the low-level information into the high-level feature representations. The cross-attention is formulated as: 
\begin{equation}
    CA(\mathbf{Q},\mathbf{K}, \mathbf{V}) = \mathbf{Q} + Softmax\left( \frac{\mathbf{Q}^{T} \cdot \mathbf{K}}{\sqrt{d}} \right) \cdot \mathbf{V} 
\end{equation}
where $\mathbf{Q} \in \mathbb{R}^{H_{1}W_{1} \times C_{1}}$, $\mathbf{K} = \mathbf{V} \in \mathbb{R}^{H_{1}W_{1} \times C_{2}}$ denote the query, key, and value inputs of the attention module, $C_1$ and $C_2$ denote the channels of the inputs, $H_1$ and $W_1$ denote the height and width of the input, respectively. $d$ denotes a hyper-parameter and is set to 64. In CSFM, $\mathbf{I}_{i}^{q}$ is reshaped as both the key and value inputs while each layer of $i$-th scale high-level feature representation is reshaped as the query input of the cross-attention module. 

MCFM can be seen as a residual connection branch for the feature extractor that fuses the low-level information into the high-level representations, making up for the defects that the previous works neglect the low-level information during the query feature extraction.

\begin{figure}[t]
  \centering
   \includegraphics[width=1\linewidth]{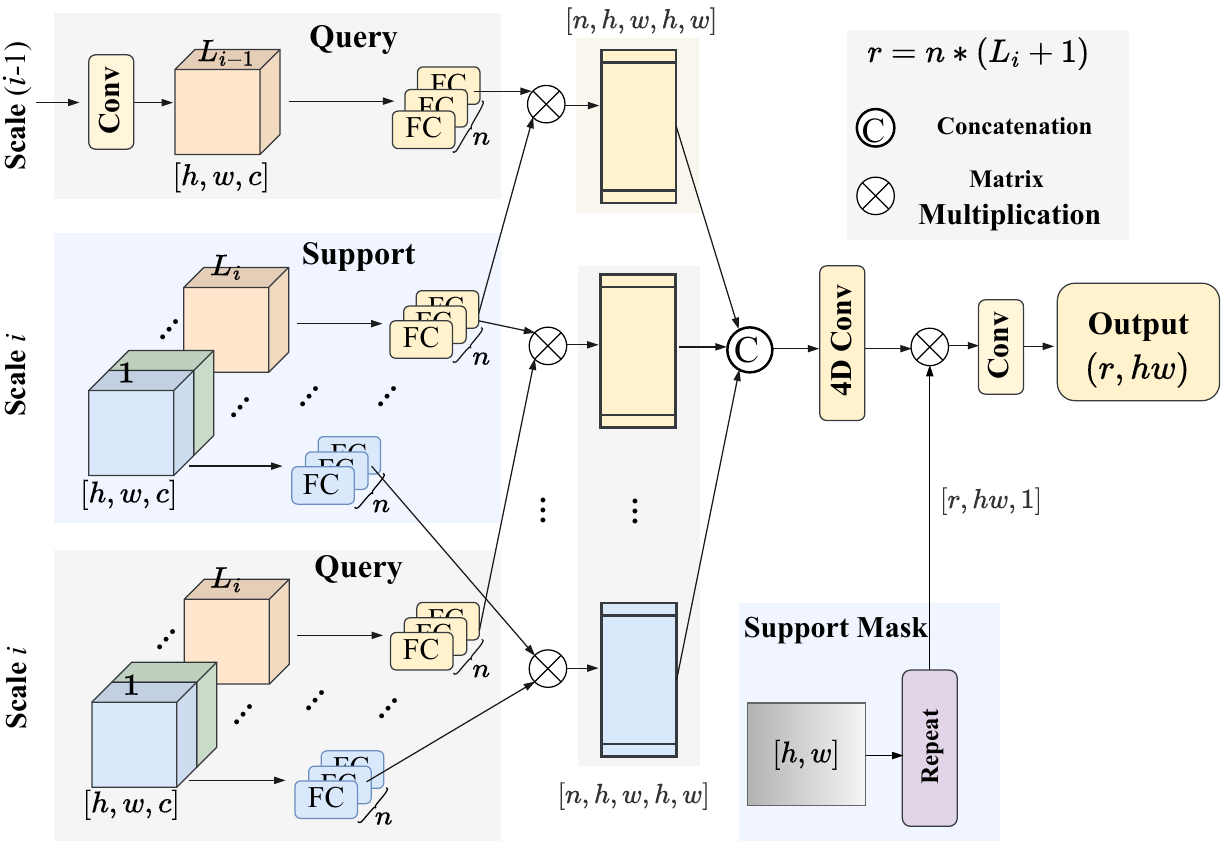}
   \caption{Illustration of Multi-Layer Interaction Module (MLIM) under the 1-Shot segmentation task. `FC' is short for fully connected layer. $n$ denotes the number of the heads. The support masks are downsampled to the corresponding size when needed.}
   \label{fig:msim}
\end{figure}

\subsection{Multi-Layer Interaction Module}
\label{msim}

To enhance the support-query correlations, we propose to interact the support-query feature representations from the layers in both the same-scale and adjacent-scale blocks of CNNs with a Multi-Layer Interaction Module (MLIM). As shown in \cref{fig:msim}, given the $i$-th scale\footnote{In this work, $i$ is either the last or the penultimate scale.} and $(i-1)$-th scale feature maps, MLIM performs same-layer mixing separately on the support-query pairs $(F_{i,l}^s, F_{i,l}^q)$ for all the layers of $i$-th scale and the adjacent-scale mixing on the support-query pairs $(F_{i,L_i}^s, F_{i-1,L_{i-1}}^q)$ for both the last layers of the adjacent scales.

Specifically, we conduct the same-layer mixing by first respectively inputting the support-query pairs to multi-head projectors and then performing an inner product to the outputs of multi-head projectors to obtain the same-scale same-layer correlations. Each head projector is a fully-connected layer. Similarly, the adjacent-scale mixing is conducted by first inputting the support-query pairs $(F_{i,L_i}^s, F_{i-1,L_{i-1}}^q)$ into multi-head projectors and then obtaining the adjacent-scale correlations with an inner product operation. Note that $F_{i-1,L_{i-1}}^q$ requires to be scaled to the same size as  $F_{i,L_i}^s$ with a convolutional operation. Then, both the same-scale same-layer correlations and the adjacent-scale correlations are concatenated into a fusion representation, followed by a series of convolutional layers to enhance the foreground-background interactions between the support-query pairs. Lastly, we weigh the support-query correlation with the support masks (downsampled when needed) to aggregate the information for the query mask prediction.

Due to the huge difference between support and query samples, the existing methods focus on the same-layer similarity, neglecting the shift issue between different layers and scales. Our method is the first work to introduce adjacent-layer query-support correspondence, which remedies the inconsistency between adjacent layers.

\subsection{Multi-Scale Mask Prediction}

After obtaining the support-query correlations, the query mask prediction is required to restore the correlations to the same size as the input query images. However, the significant size differences between the support-query correlations and the query images inevitably result in information loss. To address this issue, we propose to segment the query images with a multi-scale mask prediction strategy, where each scale enjoys a branch, and the information is comprehensively interacted from different branches in a bidirectional way. In this work, we only consider two branches, $ie$, the input-size branch and $\frac{1}{4}$ input-size branch.

Given the support-query correlations at two scales, we use them together to restore the query mask at two different scales. To improve the small-scale mask prediction, we fuse the large-scale support-query correlation (downsampled when needed) into the small-scale restoration branch. As feedback, the small-scale correlation is fused back to the large-scale restoration branch (upsampled when needed) after fusing the information from the large-scale restoration branch with a series of convolutional operations. Finally, the small-scale predicted mask is taken as a feature representation to fuse the large-scale restoration branch for the large-scale mask prediction. In such a bidirectional fusion way between the two scales, the contextual information has comprehensively interacted.

To remedy the information loss, we apply the skip connection strategy to fuse the different scale feature representations into both the small-scale and large-scale restoration branches (upsampled when needed) via concatenation, as illustrated in \cref{fig:framework}. Note that only the foreground support features at different scales are concatenated. After fusing different scale features, both branches predict the masks with two separate classifier heads, each containing a series of convolutional and ReLU operations. The difference between two classifier heads is that there are two upsampling operations and an Atrous Spatial Pyramid Pooling (ASPP) module \cite{deeplabv3+} between two convolutional blocks in the large-scale prediction line. Each convolutional block consists of two convolutional layers connected by a ReLU operation.


\subsection{Training Loss}

We use two binary cross-entropy (\textbf{BCE}) losses to respectively supervise the training of the small-scale mask prediction and the final mask prediction, composing the final target segmentation loss $\mathcal{L}_T$.
\begin{equation}\label{equ:objective}
   \mathcal{L}_T = \textbf{BCE}(y, M) + \lambda ~\textbf{BCE}(y_{small}, M_{small})
\end{equation}
where $\lambda$ is a hyper-parameter to balance the two items. $y_{small}$ and $y$ denote the small-scale and original-scale mask predictions of the query image, respectively. $M_{small}$ and $M$ denote the corresponding small-scale and original-scale ground-truth masks, respectively.

\begin{figure}[t]
  \centering
  \setlength{\abovecaptionskip}{0.4cm}
  \begin{overpic}[width=1.0\linewidth]{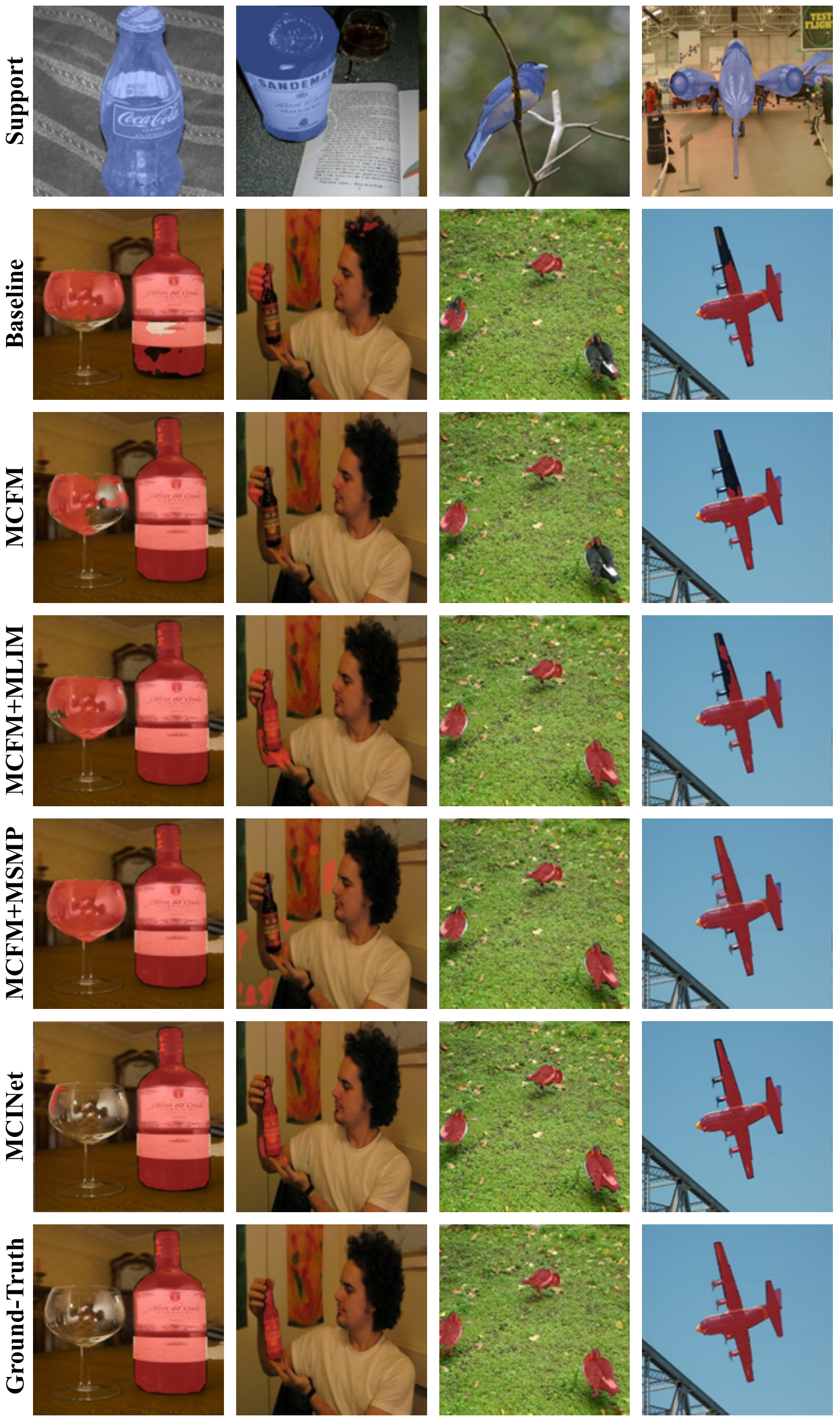}
  \small
  \put(8,-2){(a)}
   \put(22,-2){(b)}
   \put(36,-2){(c)}
   \put(51,-2){(d)}
  \end{overpic}
   \caption{Visualization of predicted results of our approach with different modules under the 1-Shot task with ResNet50 backbone on the PASCAL dataset.}
   \label{fig:ablation_blocks}
\end{figure}
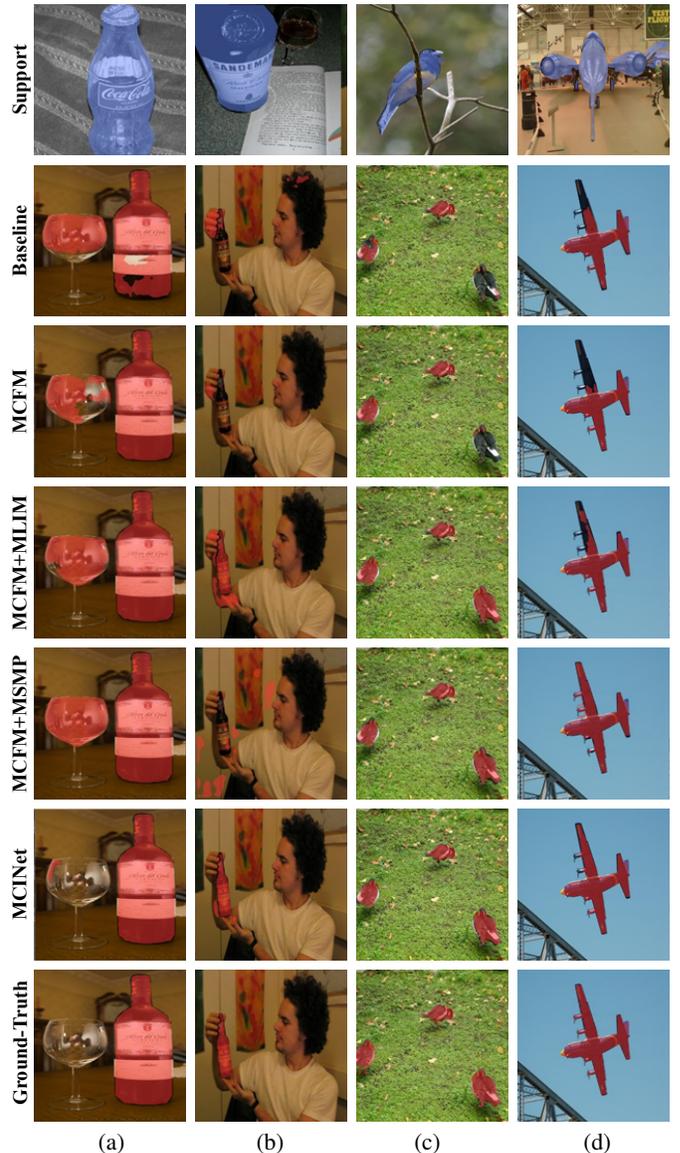

\begin{table*}[t]
  \centering
  \makebox[\textwidth]{
  \resizebox{\linewidth}{!}{
    \begin{tabular}{l@{\hspace{0.05cm}}|l|c|cccc|c|c|cccc|c|c}
    \toprule
    \multicolumn{15}{c}{PASCAL-5$^i$} \\
    \midrule
    \multicolumn{1}{l@{\hspace{0.05cm}}|}{\multirow{2}[4]{*}{Backbone}} & \multirow{2}[4]{*}{Methods} & \multicolumn{1}{c|}{\multirow{2}[4]{*}{Type}} & \multicolumn{6}{c|}{1-Shot}           & \multicolumn{6}{c}{5-Shot} \\
\cline{4-15}          & \multicolumn{1}{c|}{} &       & \multicolumn{1}{p{2.75em}}{Fold-0} & \multicolumn{1}{p{2.75em}}{Fold-1} & \multicolumn{1}{p{2.75em}}{Fold-2} & \multicolumn{1}{p{2.75em}}{Fold-3} & \multicolumn{1}{p{2.5em}|}{mIoU} & FB-IoU & \multicolumn{1}{p{2.75em}}{Fold-0} & \multicolumn{1}{p{2.75em}}{Fold-1} & \multicolumn{1}{p{2.75em}}{Fold-2} & \multicolumn{1}{p{2.75em}}{Fold-3} & \multicolumn{1}{p{2.5em}|}{mIoU} & FB-IoU \\
    \midrule
          & PPNet (ECCV'20)\cite{ppnet} &       & 48.6  & 60.6  & 55.7  & 46.5  & 52.8  & - & 58.9  & 68.3  & 66.8 & 58.0  & 63.0  & - \\
          & RPMM (ECCV'20)\cite{pmm} &       & 55.2  & 66.9  & 52.6  & 50.7  & 56.3  & - & 56.3  & 67.3  & 54.5  & 51.0  & 57.3  & - \\
          & RePRI (CVPR'21)\cite{repri} &  & 60.2  & 67.0  & \underline{61.7}  & 47.5  & 59.1  & - & 64.5  & 70.8  & \underline{71.7}  & 60.3  & 66.8  & - \\
          & CWT (ICCV'21)\cite{cwt} &       & 56.3  & 62.0  & 59.9  & 47.2  & 56.4  & -  & 61.3  & 68.5  & 68.5  & 56.6  & 63.7  & - \\
          & CAPL (CVPR'22)\cite{tian2022capl} &  Prototype      & -  & -  & -  & -  & 62.2  & -  & -  & -  & -  & -  & 67.1  & - \\
     & DCP (IJCAI'22)\cite{dcp} &       & 63.8 & 70.5 & 61.2 & 55.7 & 62.8  & - & 67.2 & 73.2 & 66.4 & 64.5 & 67.8  & - \\
          & NTRENet (CVPR'22)\cite{ntrenet} &       & 65.4  & 72.3  & 59.4  & 59.8  & 64.2  & \underline{77.0}  & 66.2  & 72.8  & 61.7  & 62.2  & 65.7  & 78.4 \\
         ResNet-50 & AAFormer (ECCV'22)\cite{aaformer} &       & \underline{69.1}  & \textbf{73.3}  & 59.1  & 59.2  & \underline{65.2}  & 73.8  & \underline{72.5}  & \textbf{74.7}  & 62.0  & 61.3  & 67.6  & 76.2  \\
          & SSP (ECCV'22)\cite{ssp} &       & 60.5  & 67.8  & \textbf{66.4}  & 51.0  & 61.4  & -  & 67.5  & 72.3  & \textbf{75.2}  & 62.1  & 69.3  & -  \\
\cline{2-15}          & CyCTR (NIPS'21)\cite{cyctr} &       & 65.7  & 71.0  & 59.5  & 59.7  & 64.0  & -  & 69.3  & 73.5  & 63.8  & 63.5  & 67.5  & -  \\
        & HSNet (ICCV'21)\cite{hsnet} &       & 64.3  & 70.7  & 60.3  & \underline{60.5}  & 64.0  & 76.7  & 70.3  & 73.2  & 67.4  & \underline{67.1}  & \underline{69.5}  & \underline{80.6} \\
          & DCAMA (ECCV'22)\cite{dcama} & Pixelwise & 67.5  & 72.3  & 59.6  & 59.0  & 64.6  & 75.7  & 70.5  & 73.9  & 63.7  & 65.8  & 68.5  & 79.5 \\
          & CMNet (TMM'22)\cite{cmnet} &       & 65.4  & 71.5  & 55.2 & 58.1  & 62.5  & 73.5  & 67.0  & 71.7  & 55.8  & 59.9  & 63.6  & 74.1 \\
          & MCINet (Ours)  &       & \textbf{69.7} & \underline{73.1} & 61.3  & \textbf{61.8} & \textbf{66.5} & \textbf{78.5} & \textbf{73.9} & \underline{74.0} & 65.7  & \textbf{68.0} & \textbf{70.4} & \textbf{81.2} \\  
          \midrule

          & PPNet (ECCV'20)\cite{ppnet}  & & 52.7  & 62.8  & 57.4  & 47.7  & 55.2  & - & 60.3  & 70.0  & \underline{69.4} & 60.7  & 65.1  & - \\
          & RePRI (CVPR'21)\cite{repri}  & & 59.6  & 68.6  & 62.2  & 47.2  & 59.4  & - & 66.2  & 71.4  & 67.0  & 57.7  & 65.6  & - \\
          & CWT (ICCV'21)\cite{cwt}  & & 56.9  & 65.2  & 61.2  & 48.8  & 58.0  & - & 62.6  & 70.2  & 68.8 & 57.2  & 64.7  & - \\
          & CAPL (CVPR'22)\cite{tian2022capl} &  Prototype     & -  & -  & -  & -  & 63.6  & -  & -  & -  & -  & -  & 68.9  & - \\
          & NTRENet (CVPR'22)\cite{ntrenet} &        & 65.5  & 71.8  & 59.1  & 58.3  & 63.7  & 75.3  & 67.9  & 73.2  & 60.1  & 66.8  & 67.0  & 78.2 \\
          & AAFormer (ECCV'22)\cite{aaformer} &       & \underline{69.9}  & \textbf{73.6}  & 57.9  & 59.7  & 65.3  & 74.9  & \textbf{75.0}  & \underline{75.1}  & 59.0  & 63.2  & 68.1  & 77.3 \\
        ResNet-101 & SSP (ECCV'22)\cite{ssp} &       & 63.2  & 70.4  & \textbf{68.5}  & 56.3  & 64.6  & -  & 70.5  & \textbf{76.4}  & \textbf{79.0}  & 66.4  & \textbf{73.1}  & - \\
\cline{2-15}          & CyCTR (NIPS'21)\cite{cyctr} &  & 67.2  & 71.7 & 57.6  & 59.0  & 63.7  & - & 71.0  & 75.0  & 58.5  & 65.0  & 67.4  & - \\
          & HSNet (ICCV'21)\cite{hsnet} & Pixelwise      & 67.3  & 72.3  & 62.0  & \underline{63.1}  & \underline{66.2}  & \underline{77.6}  & 71.8  & 74.4  & 67.0  & \underline{68.3}  & 70.4  & 80.6 \\
          & DCAMA (ECCV'22)\cite{dcama} &       & 65.4  & 71.4  & \underline{63.2}  & 58.3  & 64.6  & \underline{77.6}  & 70.7  & 73.7  & 66.8  & 61.9  & 68.3  & \textbf{80.8} \\
          & MCINet (Ours)   &       & \textbf{70.3}  & \underline{73.4}  & 62.6  & \textbf{63.5}  & \textbf{67.5}  & \textbf{78.9}  & \underline{73.3}  & 75.0  & 64.7  & \textbf{69.1}  & \underline{70.5}  & \underline{80.7} \\

    \midrule
    \multicolumn{15}{c}{COCO-20$^i$} \\
    \midrule
          & PPNet (ECCV'20)\cite{ppnet} &       & 36.5  & 26.5  & 26.0  & 19.7  & 27.2  & - & 48.9  & 31.4  & 36.0  & 30.6  & 36.7  & - \\
          & RPMM (ECCV'20)\cite{pmm} &  & 29.5  & 36.8  & 28.9  & 27.0  & 30.6  & - & 33.8  & 42.0  & 33.0  & 33.3  & 35.5  & - \\
          & RePRI (CVPR'21)\cite{repri} &       & 31.2  & 38.1  & 33.3  & 33.0  & 34.0  & - & 38.5  & 46.2  & 40.0  & 43.6  & 42.1  & - \\
        & CWT (ICCV'21)\cite{cwt} &    & 32.2  & 36.0  & 31.6  & 31.6  & 32.9  & - & 40.1  & 43.8  & 39.0  & 42.4  & 41.3  & - \\
          & CAPL (CVPR'22)\cite{tian2022capl} &  Prototype     & -  & -  & -  & -  & 39.8  & -  & -  & -  & -  & -  & \underline{48.3}  & - \\
          & DCP (IJCAI'22)\cite{dcp} &       & 40.9 & 43.8 & \underline{42.6}  & 38.3 & 41.4 & - & 45.8 & 49.7 & 43.7 & 46.6 & 46.5 &  \\
       & NTRENet (CVPR'22)\cite{ntrenet} &       & 36.8  & 42.6  & 39.9  & 37.9  & 39.3  & 68.5  & 38.2  & 44.1  & 40.4  & 38.4  & 40.3  & 69.2 \\
    ResNet-50 & AAFormer (ECCV'22)\cite{aaformer} &       & 39.8  & 44.6  & 40.6  & 41.4  & 41.6  & 67.7  & 42.9  & 50.1  & 45.5  & \underline{49.2}  & 46.9  & 68.2  \\
        & SSP (ECCV'22)\cite{ssp} &       & 35.5  & 39.6  & 37.9  & 36.7  & 37.4  & -  & 40.6  & 47.0  & 45.1  & 43.9  & 44.1 & - \\
\cline{2-15}          & CyCTR (NIPS'21)\cite{cyctr} & & 38.9  & 43.0  & 39.6  & 39.8  & 40.3  & - & 41.1  & 48.9  & 45.2  & 47.0  & 45.6  & - \\
          & HSNet (ICCV'21)\cite{hsnet} &       & 36.3  & 43.1  & 38.7  & 38.7  & 39.2  & 68.2  & 43.3  & \underline{51.3}  & 48.2  & 45.0  & 46.9  & 70.7 \\
          & DCAMA (ECCV'22)\cite{dcama} &  Pixelwise & 41.9  & \underline{45.1}  & \textbf{44.4}  & \underline{41.7}  & \underline{43.3}  & \underline{69.5} & 45.9  & 50.5  & \underline{50.7}  & 46.0  & \underline{48.3}  & \underline{71.7} \\
          & CMNet (TMM'22)\cite{cmnet} &       & \textbf{48.7}  & 33.3  & 26.8  & 31.2  & 35.0  & -  & \underline{49.5}  & 35.6  & 31.8  & 33.1  & 37.5  & - \\
          & MCINet (Ours)  &       & \underline{42.8} & \textbf{47.5} & \textbf{44.4} & \textbf{44.5} & \textbf{44.8} & \textbf{69.7}  & \textbf{50.6} & \textbf{54.4} & \textbf{55.0} & \textbf{53.0} & \textbf{53.3} & \textbf{74.8} \\ 
          \midrule

          & CWT (ICCV'21)\cite{cwt} &       & 30.3  & 36.6  & 30.5  & 32.2  & 32.4  & -& 38.5  & 46.7  & 39.4  & 43.2  & 42.0  & - \\
          & CAPL (CVPR'22)\cite{tian2022capl} &       & -  & -  & -  & -  & 42.8  & -  & -  & -  & -  & -  & 50.4  & - \\
       & NTRENet (CVPR'22)\cite{ntrenet} &  Prototype     & 38.3  & 40.4  & 39.5  & 38.1  & 39.1  & 67.5  & 42.3  & 44.4  & 44.2  & 41.7  & 43.2  & 69.6 \\
        ResNet-101 & SSP (ECCV'22)\cite{ssp} &       & 39.1  & 45.1  & 42.7  & 41.2  & 42.0  & -  & 47.4  & 54.5  & 50.4  & \underline{49.6}  & 50.2  & - \\
\cline{2-15}           & HSNet (ICCV'21)\cite{hsnet} &  & 37.2  & 44.1  & 42.4  & \underline{41.3}  & 41.2  & 69.1  & 45.9  & 53.0  & 51.8  & 47.1  & 49.5  & 72.4 \\
          & DCAMA (ECCV'22)\cite{dcama} & Pixelwise   & \underline{41.5}  & \underline{46.2}  & \underline{45.2}  & \underline{41.3}  & \underline{43.5}  & \underline{69.9}  & \underline{48.0}  & \textbf{58.0}  & \textbf{54.3}  & 47.1  & \underline{51.9}  & \underline{73.3} \\
            & MCINet (Ours)  &     & \textbf{45.5} & \textbf{50.5} & \textbf{45.7} & \textbf{43.9} & \textbf{46.4} & \textbf{70.5}  & \textbf{50.8} & \underline{55.0}  & \underline{53.7}  & \textbf{52.8} & \textbf{53.1} & \textbf{73.5} \\
    \bottomrule
    \end{tabular}}}%
    \caption{FSS performances (\%) on PASCAL-5$^i$ and COCO-20$^i$ with different backbones (ResNet50 and ResNet101). The results of all the competitors are from the published literature. The best and the second best results are marked in \textbf{bold} and \underline{underline}, respectively. }
  \label{tab:SoT}%
\end{table*}%


\begin{table}[t]
  \centering
  \small
    \begin{tabular}{l@{\hspace{0.01cm}}|l@{\hspace{0.05cm}}|c@{\hspace{0.1cm}}c@{\hspace{0.1cm}}|c@{\hspace{0.1cm}}c@{\hspace{0.1cm}}}
    \hline
    \multirow{2}[4]{*}{backbone} & \multirow{2}[4]{*}{methods} & \multicolumn{2}{c|}{1-Shot} & \multicolumn{2}{c}{5-Shot} \bigstrut\\
\cline{3-6}          &       & mIoU  & FB-IoU & mIoU  & FB-IoU \bigstrut\\
    \hline
    \multirow{4}[2]{*}{ResNet50} & HSNet\cite{hsnet} & 85.5  & -     & 86.5  & - \bigstrut[t]\\
          & DCAMA\cite{dcama} & 88.2  & 92.5  & 88.8  & 92.9 \\
          & CMNet\cite{cmnet} & 82.5  & -     & 83.8  & - \\
          & MCINet (Ours)  & \textbf{88.9}  & \textbf{93}    & \textbf{89.8}  & \textbf{93.7} \bigstrut[b]\\
    \hline
    \multirow{4}[2]{*}{ResNet101} & DAN\cite{dan}   & 85.2  & -     & 88.1  & - \bigstrut[t]\\
          & HSNet\cite{hsnet} & 86.5  & -     & 88.5  & - \\
          & DCAMA\cite{dcama} & 88.3  & 92.4  & 89.1  & 93.1 \\
          & MCINet (Ours)  & \textbf{89.3}  & \textbf{93.2}  & \textbf{89.7}  & \textbf{93.5} \bigstrut[b]\\
    \hline
    \end{tabular}%
    \caption{FSS performances (\%) on FSS-1000 with different backbones (ResNet50 and ResNet101). The results of all the competitors are from the published literature. The best results are marked in \textbf{bold}.}
  \label{tab:fss-1000}%
\end{table}%

\section{Experiments}
\label{sec:Experiments}

\textbf{Datasets.} We conduct experiments on three popular FSS benchmarks, $ie$, PASCAL-5$^i$ \cite{everingham2010pascal}, COCO-20$^i$ \cite{lin2014microsoft} (short for PASCAL and COCO, respectively), and FSS-1000 \cite{li2020fss}. Following the previous works \cite{oslsm,ppnet,panet}, we split PASCAL and COCO datasets into four folds to evaluate the model, where three folds are used for training the model and the remaining one for evaluation. Specifically, each fold of the COCO (PASCAL) dataset consists of 20 (5) categories. Compared with PASCAL, COCO is a more challenging dataset as each image in COCO usually contains multiple objects against a complex background. During the inference stage, we randomly sample 1,000 FSS tasks to evaluate the model on both PASCAL and COCO datasets. Following \cite{li2020fss}, we divide FSS-1000 dataset, which contains 1000 classes, into training set, validation set, and testing set for 520, 240, and 240 classes, respectively. The testing set contains 2,400 episodes for evaluation. The popular mean Intersection-over-Union (mIoU, the higher means the better) and Foreground-Background Intersection-over-Union (FB-IoU, higher is better) are applied as the evaluation metrics for both 1-Shot and 5-Shot segmentation tasks. If not specific, mIoU is taken as the default metric for evaluation. 

\textbf{Implementation Details.} We conduct experiments on two widely backbones, $ie$, ResNet50 and ResNet101 \cite{he2016deep}. Following the previous works, the backbones are pre-trained on ImageNet and frozen during the training stage. We apply the SGD optimizer with 0.001 learning rate and 0.9 momentum to train the model. We train 100 epochs and 50 epochs for PASCAL and COCO datasets, respectively. The batch size is set to 24 for both datasets. The images from both datasets are resized to 384 $\times$ 384. The hyper-parameter $\lambda$ is set to 0.6 for the 1-Shot task and 1.0 for the 5-Shot task on both datasets. Our model is trained on the PyTorch with two NVIDIA Tesla A100 GPUs. The codes would be released.

\begin{table}[t]
  \centering
    \small
    \begin{tabular}{c@{\hspace{0.1cm}}c@{\hspace{0.1cm}}c|c@{\hspace{0.1cm}}c@{\hspace{0.1cm}}c@{\hspace{0.1cm}}c|l}
    \toprule
  MCFM & MLIM & MSMP & {Fold0} & Fold1 & {Fold2} & {Fold3} & mIoU   \\
    \midrule
    \multicolumn{1}{c}{} &  &  & 67.5  & 72.2  & 60.6  & 56.5  & 64.2 \\
    $\checkmark$     &  &  & 68.7  & 72.6  & 60.7  & 60.4  & 65.6 \textbf{(+1.4)}  \\
      & $\checkmark$      &  & 68.6  & 72.6  & 61.7  & 56.6  &  64.9 \textbf{(+0.7)}  \\
      &  & $\checkmark$      & 69.6  & 72.3  & 57.4  & 59.6  & 64.7 \textbf{(+0.5)} \\
    $\checkmark$     & $\checkmark$     & & 70.0  & 73.0  & 60.1  & 60.2  & 65.8 \textbf{(+1.6)} \\
    $\checkmark$     &   & $\checkmark$     & 69.5  & 72.3  & 59.6  & 61.7  & 65.8 \textbf{(+1.6)}  \\
     & $\checkmark$     & $\checkmark$     & 70.6  & 72.8  & 59.9  & 61.1  & 66.1 \textbf{(+1.9)}  \\
    $\checkmark$     & $\checkmark$     & $\checkmark$     & 69.7  & 73.1  & 61.3  & 61.8  & 66.5 \textbf{(+2.3)} \\
    \bottomrule
    \end{tabular}%
  \caption{The ablation results (\%) of our MCINet on the PASCAL dataset with the ResNet50 backbone for the 1-Shot segmentation task. }
\label{tab:ablation}%
\end{table}%

\begin{figure}[t]
  \centering
   \includegraphics[width=0.9\linewidth]{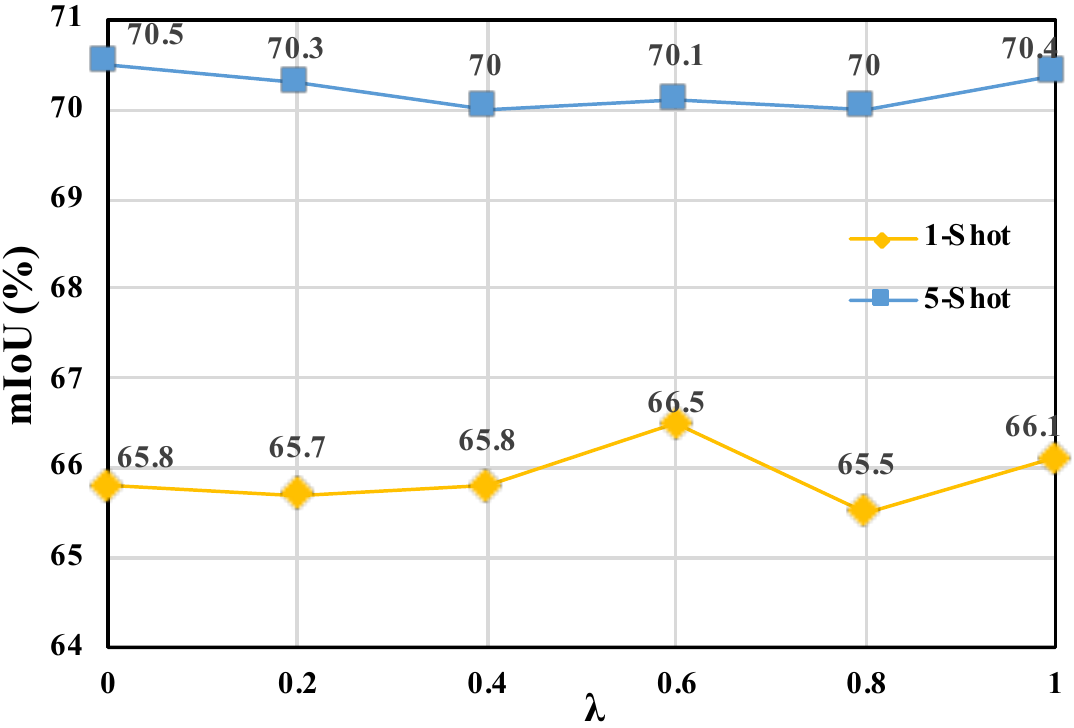}
   \caption{Impacts of hyper-parameter $\lambda$ under both 1-Shot and 5-Shot tasks on the PASCAL dataset with ResNet50 backbone. }
   \label{fig:lambda}
\end{figure}

\subsection{Comparison with State-of-the-Art}
To evaluate the effectiveness of the proposed model, we conduct extensive comparisons with the state-of-the-art (SOTA) competitors with two backbones under two few-shot settings. \cref{tab:SoT} demonstrates the quantitative comparison results on the two benchmarks. From the results, we observe that the proposed MCINet performs very competitively on both two benchmarks. For the PASCAL dataset, MCINet achieves 66.5\% and 67.5\% with ResNet50 and ResNet101 backbones and substantially outperforms the second-best competitors with both 1.3\% improvements under the 1-Shot segmentation task. For the 5-Shot segmentation task, the proposed MCINet achieves 70.4\% with ResNet50 and outperforms the second-best competitors with 0.9\%. For the stronger ResNet101, MCINet performs very competitively for Fold-0, Fold-1, and Fold-3 but is slightly inferior for Fold-2, and thus performs the second best under the overall mIoU metric. For the more challenging COCO dataset, the proposed MCINet obtains the best performances on both mIoU and FB-IoU metrics for both 1-Shot and 5-Shot segmentation tasks with ResNet50 and ResNet101 backbones. Specifically, MCINet achieves 44.8\% and 46.4\% respectively with ResNet50 and ResNet101 for the 1-Shot segmentation task, which outperforms the second-best competitors by 1.5\% and 2.9\%. In the 5-Shot task, MCINet promotes the SOTA performance up to 53.3\% with ResNet50, which significantly surpasses the second-best competitor by 5.0\%. When using the ResNet101 backbone, our approach obtains 53.1\%, beating the follower by 1.2\%. For the FB-IoU metric, MCINet achieves 70.5\% with ResNet101 for the 1-Shot and 74.8\% with ResNet50 for the 5-Shot, respectively achieving the best performances under both tasks.

As shown in \cref{tab:fss-1000}, our MCINet reaches SOTA in both mIoU and FB-IoU metrics for both 1-Shot and 5-Shot segmentation tasks with ResNet50 and ResNet101 backbones. For mIoU metric, MCINet achieves 88.9\% and 89.3\% with ResNet50 and ResNet101 backbones in the 1-Shot setting, and 89.8\% and 89.7\% with ResNet50 and ResNet101 backbones in the 5-Shot setting, respectively. For FB-IoU metric, MCINet also performs well, achieving 93.2\% with ResNet50 in 1-Shot segmentation task and 93.7\% with ResNet101 in the 5-Shot segmentation task. Those promising results demonstrate the effectiveness of MCINet. 

\subsection{Ablation Study}

In this subsection, we design a series of ablation studies to evaluate the effects of different modules. All the results are obtained with the ResNet50 backbone under the 1-Shot task on the PASCAL dataset. 

As shown in \cref{tab:ablation}, MCFM obtains performance gain for all Folds and significantly improves the overall performance by 1.4\%, which indicates the effectiveness of low-level structural information for the segmentation tasks. Besides, both MLIM and MSMP could separately improve the baseline for the overall performance, indicating the effectiveness of each module. When two of the three modules are equipped together, the performances are further improved, especially for Fold-0 and Fold-3. When integrating all the three modules together, our method gains an extra 0.9\% overall performance against the model only with MCFM and improves the baseline from 64.2\% to 66.5\% on the overall mIoU metric. The effects of each module are more obvious on the COCO dataset under the 5-Shot task and are dropped due to the space limit.

To better show the effects of each module, we visualize the predictions of some 1-Shot segmentation tasks from the PASCAL dataset with the ResNet50 backbone in \cref{fig:ablation_blocks}. From the results, we observe that all the modules could bring positive effects for the segmentation results and the model integrating with all the modules could significantly improve the predictions via eliminating distractions ((a) in \cref{fig:ablation_blocks}), remaining precise details (b), perceiving multiple objects in an image (c), and segmenting the complete object (d).

\textbf{Low-level Feature.} In MCFM, the low-level structural information from another query branch is fused with high-level information. As shown in \cref{fig:low_feature}, we compare the low-level and high-level features from the same backbone with our method in PASCAL dataset under 1-Shot setting with ResNet50 backbone. From the results, we observe that our method is more balance in four folds and achieves the best performance. Fused with the low-level feature from the same backbone brings a slight improvement, compared with none feature supplementary.

\textbf{Hyper-parameter $\lambda$.} \cref{fig:lambda} illustrates the impacts of the hyper-parameter $\lambda$ to the overall mIoU on the PASCAL dataset. We observe that the performance of 1-Shot task is sensitive to $\lambda$ and the model obtains the best when $\lambda$ equals 0.6. However, the 5-Shot performance gets slightly worse when $\lambda$ is nonzero, indicating that the low-scale supervision brings littler impacts for the 5-Shot task.

\textbf{Heads $n$.} In MLIM, the number of heads $n$ in the projectors is a hyper-parameter. In this experiment, we evaluate its impacts on the performance. \cref{tab:multi-head} illustrates the impacts of $n$ to the overall mIoU on the PASCAL dataset. Specifically, we set $n$ from $\{1,2,4,8\}$. We observe that the performance improves as $n$ increases, with the best result obtained when $n$ equals 4, and no improvement obtained when $n$ is larger than 4. For the computational complexity, we set $n=4$ in MLIM.

\textbf{Skip Connection.} The skip connection \cite{u-net} aims at aggregating the information of different-level features from the support-query pairs. For the support line, either the features of whole images or the features of the foreground objects are concatenated into the skip connection module. In this experiment, we evaluate the impacts of different support features in the skip connection to the overall mIoU. From the results on the PASCAL in \cref{tab:skip-connection}, we observe that the performance of the model with only the foreground support feature is better than that with the whole support feature in the skip connection.

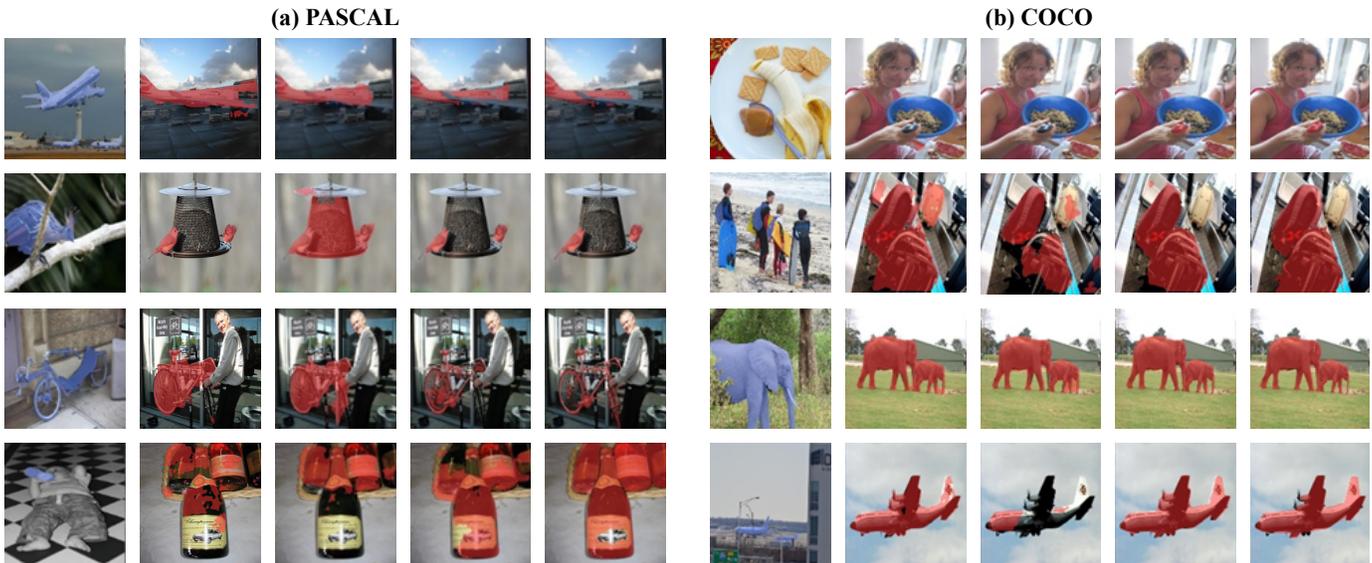
\begin{figure*}[t]
  \centering
    \setlength{\abovecaptionskip}{0.4cm}
  \begin{overpic}[width=1\linewidth]{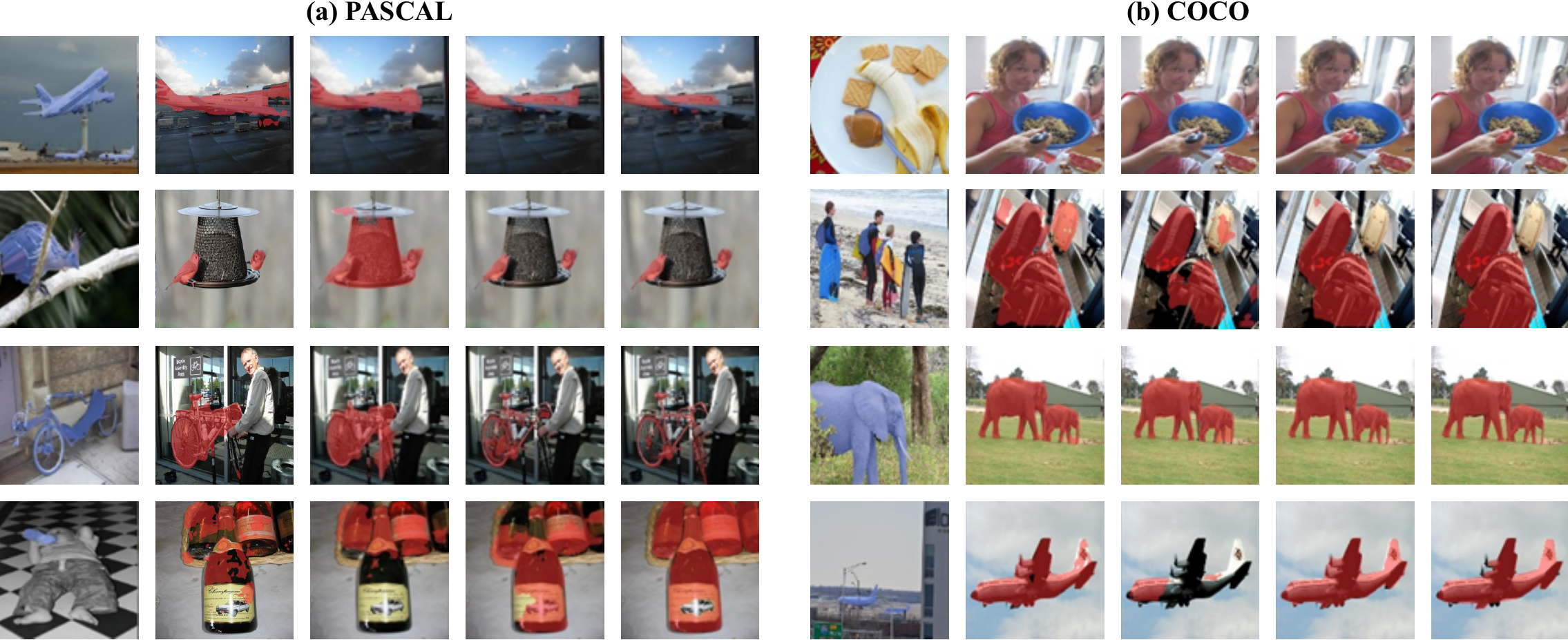}
  \put(1,-2){Support}
  \put(8,-2){DCAMA \cite{dcama}}
  \put(20,-2){HSNet \cite{hsnet}}
  \put(31,-2){MCINet}
  \put(42,-2){GT}
  \put(53,-2){Support}
  \put(60,-2){DCAMA \cite{dcama}}
  \put(72,-2){HSNet \cite{hsnet}}
  \put(83,-2){MCINet}
  \put(94,-2){GT}
    \end{overpic}
  \caption{Some qualitative 1-Shot visualization samples from both PASCAL and COCO datasets with the ResNet50 backbone for different cases. Our MCINet achieves promising mask prediction under various challenging scenarios, including small target object, object with interference, multiple target objects, and object details. `GT' denotes the query ground-truth.}
  \label{fig:coco_compare}
\end{figure*}

\begin{table}[t]
  \centering
    \begin{tabular}
    {l|c@{\hspace{0.1cm}}c@{\hspace{0.1cm}}c@{\hspace{0.1cm}}c|c}
    \hline
    Low-level Feature & Fold-0 & Fold-1 & Fold-2 & Fold-3 & Mean \bigstrut\\
    \hline
    None  & 70.6  & 72.8  & 59.9  & 61.1  & 66.1 \bigstrut[t]\\
    First Block of Backbone & 70.6  & 72.3  & 61.5  & 59.9  & 66.1 \\
    Second Block of Backbone & 70.0  & 72.7  & 60.0 & 62.1  & 66.2 \\
    Ours  & 69.7  & 73.1  & 61.3  & 61.8  & 66.5 \bigstrut[b]\\
    \hline
    \end{tabular}%
  \caption{Impacts (\%) of low-level feature in CSFM under 1-Shot task on the PASCAL dataset with ResNet50 backbone. `None' denotes removing CSFM from our method. `First Block of Backbone' and `Second Block of Backbone' denote fusing the high-level feature with the low-level feature in first or second block of the same backbone in CSFM module.}
  \label{tab:low_feature}%
\end{table}%

\begin{table}[t]
  \centering
    \begin{tabular}{c|cccc|c}
    \hline
    $n$ & Fold-0 & Fold-1 & Fold-2 & Fold-3 & \multicolumn{1}{p{2.1em}}{Mean}  \bigstrut\\
    \hline
    1     & 70.9  & 72.2  & 60.3  & 60.4  & 66.0 \\
    2     & 70.2  & 73.1  & 60.8  & 60.4  & 66.1 \\
    4     & 69.7  & 73.1  & 61.3  & 61.8  & 66.5  \\
    8     & 70.5  & 72.4  & 62.2  & 60.4  & 66.4\\
    \hline
    \end{tabular}%
    \caption{Impacts (\%) of number of heads in MSIM under 1-Shot task on the PASCAL dataset with ResNet50 backbone.}
  \label{tab:multi-head}%
\end{table}%

\begin{table}[t]
  \centering
    \begin{tabular}{l|c|c}
    \hline
    skip type & mIoU  & FB-IoU \bigstrut\\
    \hline
    whole image & 65.9  & 77.3 \bigstrut[t]\\
    foreground & 66.5  & 78.5 \bigstrut[b]\\
    \hline
    \end{tabular}%
    \caption{The ablation results (\%) of different support feature in skip connection on the PASCAL dataset with the ResNet50 backbone for the 1-Shot segmentation task. }
  \label{tab:skip-connection}%
\end{table}%


\begin{figure*}[t]
  \centering
  \includegraphics[width=1\linewidth]{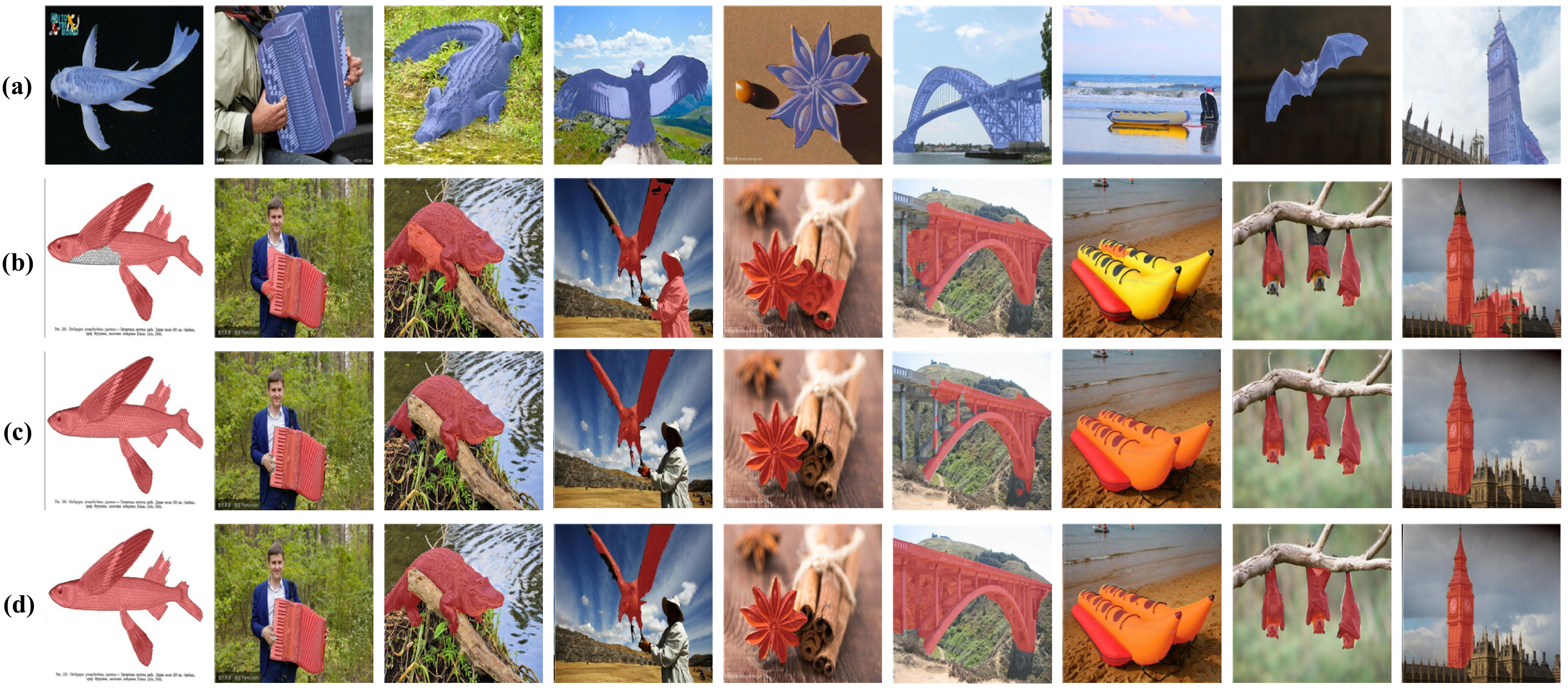}
  \caption{Some 1-Shot visualization on the FSS-1000 dataset with ResNet50 backbone. (a) support image, (b) HSNet \cite{hsnet}, (c) MCINet (Ours), (d) Ground-Truth.}
  \label{fig:fss_compare}
\end{figure*}


\subsection{Qualitative Results}

To intuitively visualize the segmentation results, we provide some qualitative results under the 1-Shot task with the ResNet50 backbone in \cref{fig:coco_compare} and \cref{fig:fss_compare} for COCO, PASCAL and FSS-1000 datasets, respectively. We also provide the comparison results of DCAMA \cite{dcama} and HSNet \cite{hsnet} with the released trained models. From the results, we observe that MCINet handles better than the competitors, especially for the cases of segmenting objects with inferences, small target objects, and multiple targets. We speculate the superior segmentation capability is that the support-query pairs are fully interacted with our proposed modules. 




\section{Conclusion}
\label{sec:conclusion}
In this work, we addressed FSS by exploiting and interacting the contextual information from same-scale, adjacent-scale, and cross-scale blocks with the proposed multi-content fusion module, multi-layer interaction module, and multi-scale mask prediction strategy. From the extensive experiments, we conclude that fusing multi-content information in query-support correlation, especially boosting the query features by incorporating the low-level structural information from another branch into the high-level representations, brings desirable bonus for performance and bidirectionally interacting the multi-scale features in mask prediction significantly boosts the FSS results. Together with the three modules, the proposed method sets new SOTA on two benchmarks.

\bibliographystyle{IEEEtran}
\bibliography{egbib}

\vfill

\end{document}